\documentclass[11pt,onecolumn]{article}

\usepackage {amssymb,rotating,graphicx,cite, calc,color,dsfont,epsfig,}
\usepackage {amssymb,rotating,graphicx,cite,
calc,color,dsfont,epsfig,amsmath,balance,multirow,multicol,mathrsfs}

\usepackage{microtype}
\usepackage{stackengine}
\usepackage[hyperref]{acl2020}
\date{}
\begin{document}
\newtheorem{example}{Example}
\newcommand\EndBox{\hspace*{\fill}~\QED\par\endtrivlist\unskip}
\title{\LARGE Evaluating Sparse Interpretable Word Embeddings for Biomedical Domain\thanks{This research was supported by the Institute for Artificial Intelligence
(iai.iut.ac.ir) under a generous grant by Mr. Farhad Rahnema.}}
\author{Mohammad Amin Samadi$^1$, Mohammad Sadegh Akhondzadeh$^1$\thanks{The first two authors contributed equally.},
Sayed Jalal Zahabi$^1$\\ Mohammad Hossein Manshaei$^1$, Zeinab Maleki$^1$, Payman Adibi$^2$\\
\\
$^1$Department of Electrical and Computer Engineering\\
Isfahan University of Technology,
Isfahan 84156-83111, Iran\\
\\
$^2$Internal Medicine Department,
Isfahan University of Medical Sciences,
Isfahan, Iran.\\
\\
\texttt{aminsamadi21@gmail.com}, \texttt{ms.akhondzadeh@ec.iut.ac.ir}\\
\texttt{\{zahabi,manshaei,zmaleki\}@iut.ac.ir}, \texttt{adibi@med.mui.ac.ir}}

\maketitle

\begin{abstract}
Word embeddings have found their way into a wide range of natural language processing tasks including those in the biomedical domain. While these vector representations successfully capture semantic and syntactic word relations, hidden patterns and trends in the data, they fail to offer interpretability. Interpretability is a key means to justification which is an integral part when it comes to biomedical applications.
We present an inclusive study on interpretability of word embeddings in the medical domain, focusing on the role of sparse methods. Qualitative and quantitative measurements and metrics for interpretability of word vector representations are provided.
For the quantitative evaluation, we introduce an extensive categorized dataset that can be used to quantify interpretability based on category theory.
Intrinsic and extrinsic evaluation of the studied methods are also presented. As for the latter, we propose datasets which can be utilized for effective extrinsic evaluation of word vectors in the biomedical domain.  Based on our experiments, it is seen that sparse word vectors show far more interpretability while preserving the performance of their original vectors in downstream tasks.
\end{abstract}

\section{Introduction}

Word vector representation algorithms are used to embed words in a high dimensional space where words maintain semantic and syntactic relationships with each other. The main idea behind most of these algorithms is to maximize the dot product of words’ vector representations appearing in the same context, and therefore, maximizing the similarity between them. These methods have been used in many down-stream Natural Language Processing (NLP) tasks and have proved to make a noticeable difference in the performance.

Although they have improved the performance in various tasks, word embedding methods fail to train word vectors with interpretable dimensions. The word relations seem relative rather than absolute and therefore can not be interpreted in order to identify what properties are being captured by these algorithms. Moreover, interpretable word vector representations are believed to capture features, similar to those typically considered in NLP. The importance of this escalates when it comes to biomedical NLP (bioNLP) applications where decisions made by machines are to be reliable, and thus interpretable. However, current methods fail to fulfill this as they work like a black-box.

One approach towards interpretability in word embedding is via sparse methods. There have been many cases (in unsupervised methods) demonstrating that interpretability is in commensurate with sparsity \cite{murphy2012learning} \cite{fyshe2014interpretable} \cite{dahiya2016discovering}.
Besides, when it comes to computation, sparse vectors are faster and easier to work with. Sparse interpretable word vectors have proved to even outperform the original vectors in some downstream tasks \cite{guo-etal-2014-revisiting}. Indeed, classifiers  benefit from higher usability of sparse representations as features and the separability of dimensions in word vectors can be profitable.

While sparse interpretable word embedding methods have been studied in the context of NLP, they are not well understood in bioNLP.
This motivates us to further explore the area of interpretability in biomedical domain, focusing mainly on sparse interpretable word vector representations.
Specifically, we first train dense word vectors on medical text, via Skip-gram \cite{mikolov2013distributed} and GloVe \cite{pennington2014glove} methods .  The obtained dense representations are then used to train sparse interpretable word embeddings through two methods, namely, Sparse Overcomplete Word Vector Representation (SPOWV) \cite{faruqui2015sparse}, and Sparse Interpretable Neural Embedding (SPINE) \cite{spine}. This way, we arrive at four different sets of sparse representations which are compared to each other, and to their original dense vectors, in terms of their performance in downstream tasks and interpretability.

Our contribution to this area is three-fold: 1) Introducing a structured pipeline for training sparse interpretable word vectors in biomedical field, and to perform appropriate assessments to evaluate them in different aspects. 2) Nominating classification tasks and datasets which can be reliable metrics as downstream tasks in the biomedical domain. 3) Proposing a novel approach for the evaluation of interpretability by using a comprehensive categorized dataset which can serve as a reference in biomedical context.\footnote{
We have made our implementations and dataset publicly available at https://github.com/Institute-for-Artificial-Intelligence/Sparse-Interpretable-Word-Embeddings-For-Medical-Domain}

The paper is organized as follows. In Section~\ref{Sec2} we review the related works to clarify where our work stands in the relevant field. In Section~\ref{Sec3}, the required preliminaries on the used dense and sparse word embeddings are briefly pointed out. Then, in Section~\ref{Sec4}, we explain our methods including the pre-processing steps, experiment setup, and the evaluation procedures. The results are then provided and discussed in Section~\ref{Sec5}. Finally, Section~\ref{Sec6} concludes the paper.
\section{Related Works}\label{Sec2}
The problem of word vector interpretability has been an active area in recent years. \cite{murphy2012learning}, for the first time, applied matrix factorization to construct a non-negative sparse embedding. Later on, \cite{faruqui2015sparse} applied an optimization problem within a dictionary learning approach to create sparse interpretable word embedding. In 2017, \cite{park2017rotated} took advantage of a rotation matrix to improve the interpretability of the word embeddings. In another recent study,  k-sparse autoencoder is utilized to create a sparse interpretable word embedding \cite{spine}.

As for the biomedical domain, most of the studies on word vector representation
merely look into the concept of similarity and the performance of word embeddings on downstream tasks \cite{sajadi2015domain} \cite{pakhomov2016corpus} \cite{chiu2018bio} \cite{muneeb2015evaluating} and the interpretability of the word embeddings is not well studied.

There is however one recent work addressing interpretability in the biomedical context \cite{jha2018interpretable} which applies a \emph{supervised non-sparse} method based on  \cite{park2017rotated}, to improve the interpretability of the word embedding.
To the best of our knowledge, there has been no prior work on \emph{unsupervised sparse} interpretable word embedding in the biomedical context.

\section{Preliminaries: Dense and Sparse Word Embeddings}\label{Sec3}
In this section, first we  point out the required preliminaries on the dense word embeddings as the baseline original representations used in this paper. Then, we briefly cover the fundamentals of the two aforementioned sparse methods, i.e., SPOWV and SPINE.
\subsection{Dense Word Embedding Methods}





Word vector representation methods have long been an integral part of NLP. These methods attempt to represent words as vectors which embody word relations from a large unlabeled corpus. They can be divided into two main categories: a) Algorithms, such as latent semantic analysis (LSA) \cite{LSA}, that use global co-occurrence information and matrix factorization, b) Algorithms like Word2Vec's Skip-gram \cite{mikolov2013distributed} that use local window-based information.

There are however, some methods such as the Global Vectors for Word Representation
(GloVe) \cite{pennington2014glove} that try to take advantage of both classes by utilizing global and window-based information together. Since Skip-gram and GloVe have shown to outperform other approaches in downstream tasks such as sentiment analysis \cite{socher2013recursive}, named entity recognition \cite{NER}, and NP parsing \cite{parsing}, in this paper, we will only use these two methods to train dense word embeddings on medical text.

Although both of these algorithms do well in capturing semantic and syntactic relations among words, they fail to train interpretable set of word vectors with meaningful individual dimensions.

\subsection{SPOWV: Sparse Overcomplete Word Vector Representation}
Faruqui et al.  proposed a novel approach in \cite{faruqui2015sparse} in order to embed distributed representation of words to a space with higher dimension, using sparse transformations. The goal was to achieve a level of interpretability that would be clear to human judgment as well. To achieve this goal, they applied sparse dictionary learning to transform the vector
$ \mathbf{X} \in \mathbb{R}^{V \times L} $
into
$ \mathbf{A} \in \mathbb{R}^{V \times K} $
where $V$ is the size of the vocabulary, $L$ is the dimension of the original space, and $K$ is the dimension of sparse space such that $K>L$. The objective function designed to optimize SPOWV is given by
\begin{equation}\label{spoweq}
\underset{\mathbf{D}, \mathbf{A}}{\arg \min } \sum_{i=1}^{V}\left\|\mathbf{X}_{i}-\mathbf{a}_{i} \mathbf{D}\right\|_{2}^{2}+\lambda\left\|\mathbf{a}_{i}\right\|_{1}+\tau\|\mathbf{D}\|_{2}^{2}
\end{equation}
where $\mathbf{D} \in \mathbb{R}^{K \times L} $ is a dictionary consisting of vector bases,  $ \mathbf{A} $ is the desired sparse embedding whose $i^{\rm th}$ row is denoted by $a_i$, and $\lambda $ and $ \tau $ are coefficients for the regularization penalty of $ \mathbf{A} $ and $ \mathbf{D}$, respectively.



\subsection{SPINE: Sparse Interpretable Neural Embedding}

The concept of sparse autoencoders were first introduced in Andrew Ng's lecture slides \cite{ng2011sparse}. SPINE \cite{spine} utilizes $k$-sparse autoencoders \cite{makhzani2013k} in order to train sparse interpretable word vector representations. Its corresponding loss function consists of three terms to induce sparsity and interpretability as follows
\begin{equation}
L(\mathcal{D})=\lambda_{1} R L(\mathcal{D})+\lambda_{2} A S L(\mathcal{D})+\lambda_{3} P S L(\mathcal{D})
\label{spine_loss}
\end{equation}
where $\mathcal{D}$ denotes the input set of representations, and $\lambda_1$, $\lambda_2$, $\lambda_3$ are parameters that should be tuned.
The terms on the right hand side of the above loss function are explained briefly as follows.
\subsubsection{Reconstruction Loss (RL)}
This term, denoted by $RL(\mathcal{D})$ in Equation (\ref{spine_loss}), minimizes the  error from reconstructing the input from the latent space which is the main loss component in non-sparse basic autoencoders as well. Considering $\widetilde{\mathbf{X}}$ to be the reconstructed vector from the input vector $\mathbf{X}$, the reconstruction loss is written as follows
\begin{equation}
R L(\mathcal{D})=\frac{1}{|\mathcal{D}|} \sum_{\mathbf{X} \in \mathcal{D}}\|\mathbf{X}-\tilde{\mathbf{X}}\|_{2}^{2}
\label{RL}
\end{equation}

\subsubsection{Average Sparsity Loss (ASL)}
This term, denoted by $ASL(\mathcal{D})$ in Equation (\ref{spine_loss}), penalizes deviations from the desired sparsity as follows
\begin{equation}
A S L(\mathcal{D})=\sum_{h \in \mathcal{H}}\left(\max \left(0, \rho_{h, \mathcal{D}}-\rho_{h, \mathcal{D}}^{*}\right)\right)^{2}
\label{spine_ASL}
\end{equation}
where
$\rho_{h, \mathcal{D}}^{*}$ is the desired sparsity ratio for unit $h$  across $\mathcal{D}$, and $\rho_{h, \mathcal{D}}$ denotes the observed average activation value  for unit $h$ across $\mathcal{D}$.

The idea behind the formulation of this penalty is that in order for the latent vector to be $k$-sparse and have $k$ active elements, the average of the sum of elements should be around
$
1 - \rho_{h, \mathcal{D}}^{*}$.

\subsubsection{Partial Sparsity Loss (PSL)}
This term, denoted by $PSL(\mathcal{D})$ in Equation (\ref{spine_loss}), penalizes the elements whose amounts are not close to 0 or 1 as follows
\begin{equation}
P S L(\mathcal{D})=\frac{1}{|\mathcal{D}|} \sum_{\mathbf{X} \in \mathcal{D}} \sum_{h \in \mathcal{H}}\left(Z_{h}^{(\mathbf{X})} \times\left(1-Z_{h}^{(\mathbf{X})}\right)\right)
\label{spine_PSL}
\end{equation}
where $Z_{h}^{(\mathbf{X})}$ denotes the activation value for the hidden unit $h$, for the input representation $\mathbf{X}$.

The role of PSL is to enforce elements of the latent space vectors to be binary rather than being around the average desired sparsity ratio.


\section{Material and Methods}\label{Sec4}
\begin{figure*}[!t]
\centering
  \includegraphics[scale=0.38]{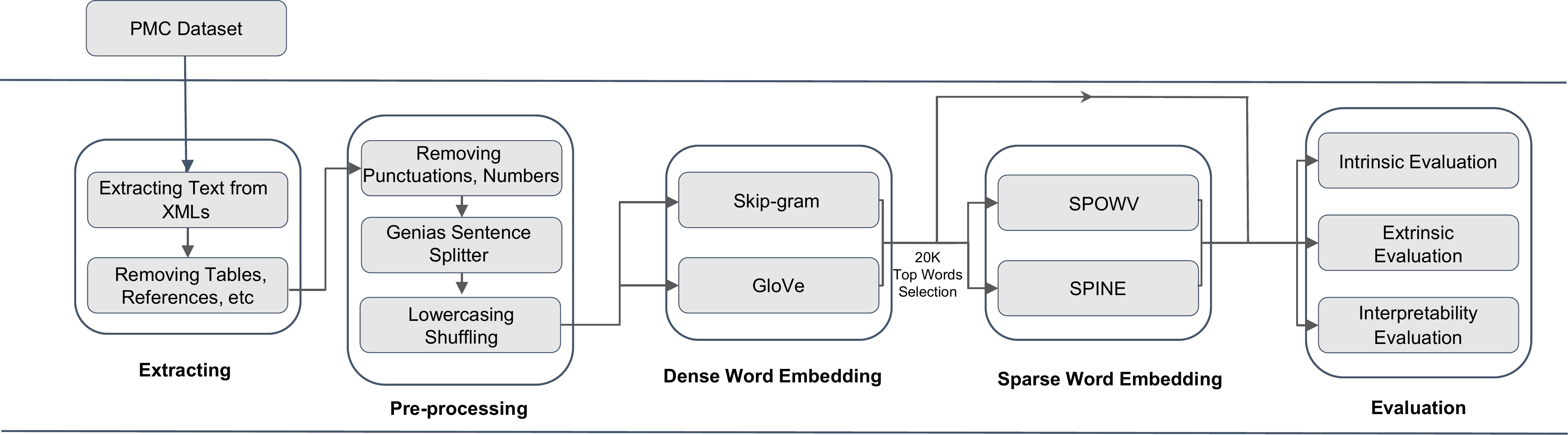}
  \caption[Methodology overview]
{Methodology overview}
  \label{fig-System}
\end{figure*}

An overview of our methodology is shown as a block diagram in Figure~\ref{fig-System}.
In this section, we explain the blocks involved in this pipeline including the pre-processing steps, experiment setup, and the evaluation procedures.
\subsection{Corpora and Pre-processing}

To form a training text for developing distributed word vector representations, we used the PubMed Central (PMC) Open Access Subset\footnote{https://www.ncbi.nlm.nih.gov/pmc/tools/openftlist/}. We believe that word vectors can benefit from domain specific training data in medical context. PMC's open access repository contains over one million free full-text digital archives of biomedical and life sciences journal literature at the U.S. National Institutes of Health's National Library of Medicine. In order to improve the quality of the trained word embeddings, we excluded non-text partitions of the papers such as tables and references from the training text of word embedding models. To achieve this, we took advantage of the structure of XML format of the dataset and a text extractor \cite{Achakulvisut2020} to extract the full body of each document to remove the undesired parts of all the papers. Then, in the pre-processing step, we first removed all the punctuation marks from the training text, replaced all the numbers with ``0", and all characters were lowercased. Finally, we utilized GENIA Sentence Splitter \cite{genias}, which is optimized for sentence segmentation of text in biomedical domain and shuffled the sentences in the end.

\subsection{Experiment Setup}

There are various factors that directly affect the dense word embedding methods. For Instance, word embeddings trained with larger window size are capable of capturing more complicated semantic relationships. In this regard, as mentioned earlier, we experiment with the two most commonly used dense word embedding methods, Skip-gram and GloVe. The training data contained over 6B tokens and the dimension size of each word vector was considered 300 with the window size of 15. Once the dense word vectors were trained on the whole training data,
we  selected 20k most frequent words as the input to the sparse methods. The sparse interpretable word vectors were trained with dimension size 1000.

\subsection{Hyperparameter Tuning}
\label{hyper}


With the various coefficients present in the loss functions of SPOWV, i.e.,
($ \lambda, \tau $) in Equation (\ref{spoweq}), and SPINE i.e.,
($ \lambda_{1}, \lambda_{2}, \lambda_{3} $) in Equation (\ref{spine_loss}), their performance in downstream tasks can not fairly represent how interpretable they are. This demands for a hyperparameter tuning. To do so, we design a task inspired by \cite{newman2010automatic} that could demonstrate interpretability in a set of word vectors. It is reasonable to expect that in an interpretable word embedding space, top words of dimensions would form a semantically coherent group. Since doing so for each dimension is resource-demanding and time-consuming, we pick words randomly from biomedical terminologies that represent specific semantic groups. For each word, we identify top words of the dimensions in which the word was active. To have a quantitative measurement, we sum up the cosine similarity between every pair of the top words in the original space. The more the total sum for the resulting vectors with a set of parameters, the more interpretable it is on the selected words representing medical semantic groups.


\subsection{Evaluation of Interpretability}

In this section, we explain the qualitative and quantitative methods used in this paper to assess the interpretability of the word embeddings under study.

\subsubsection{Qualitative Evaluation}
\label{inter_qual}

If a word vector space is interpretable, words active in a dimension should form a semantically or syntactically coherent group. In order to examine interpretability qualitatively, we pick words in the medical domain that could represent a specific medical category.
For each word we seek for the dominating dimension, i.e., the dimension which has the highest value, and will find the top five words with the highest values in that dimension. The more these words are closely related to each other and the considered word, the more interpretable the embedding space and that dimension is.


\subsubsection{Quantitative Evaluation}
\label{inter_quant}

The most common way of quantitative evaluation of interpretability has been the word intrusion task. Word intrusion task is a multiple choice question on words extracted from the word embedding dimensions. Specifically, in order to examine interpretability with word intrusion task, 4 out of 5 choices are selected from the top 10 percent of a dimension and the other choice is a word selected from the bottom half of the dimension, but in the top 20 percent of another dimension. Interpretability of a set of word vectors is measured by how well humans can detect the intruding words in the described task.

Although the word intrusion task is very effective in evaluating interpretability, since we are investigating interpretability in biomedical domain, gathering people with sufficient knowledge to perform the word intrusion task reliably proved to be challenging and time-consuming. Therefore, we borrowed the evaluation method firstly introduced in \cite{semantic_structure}. As mentioned there, a categorized dataset which is big enough can approximate the word intrusion task. In this vein, we used the UMLS Metathesaurus \cite{UMLS} that is a large dictionary for biomedical terminologies. It includes definitions, relations, semantic categorization, etc. We took advantage of the semantic grouping available in Metathesaurus, for this purpose. There are originally 127 categories of medical concepts in the dataset. Nearly half of the 20000 words in our word vectors vocabulary are present and grouped in the dataset. We discarded semantic groups with lower than 5 and higher than 250 words since the grouping could have been too specific or too general, respectively. Eventually, we ended up with 93 semantic groups containing 62 words on average. We believe the resulting dataset can be a reliable measure for interpretability in biomedical domain.

Based on this approach, a score is given to each category-dimension pair as follows.
\begin{equation}
\begin{aligned} I S_{i, j}^{+} &=\frac{\left|S_{j} \cap V_{i}^{+}\left(\gamma \times n_{j}\right)\right|}{n_{j}} \times 100 \\ I S_{i, j}^{-} &=\frac{\left|S_{j} \cap V_{i}^{-}\left(\gamma \times n_{j}\right)\right|}{n_{j}} \times 100 \end{aligned}
\end{equation}
where $ I S_{i, j}^{+} $  and $ I S_{i, j}^{-} $ are respectively, the positive and negative interpretability scores of the $i^{\rm th} $ dimension in $ j^{\rm th} $ category, $S_{j}$ is the set of words present in the $ j^{\rm th}$ category and $ n_{j} $ is the size of this set. The parameters $V_{i}^{+}\left(\gamma \times n_{j}\right)$ and $V_{i}^{-}\left(\gamma \times n_{j}\right)$ are the top and bottom $\gamma \times n_{j}$ words of the $ i^{\rm th} $ dimension, respectively, with $\gamma$ being a natural number controlling the strictness of the score.

The final interpretability score of the $(i-j)$ (dimension--category) pair is given by
\begin{equation}
IS_{i, j} =\max \left(I S_{i, j}^{+}, I S_{i, j}^{-}\right)
\end{equation}
Now, let the interpretability score of the $ i^{\rm th} $ dimension, denoted by $IS_{i}$, be the maximum score of any categories in that dimension, i.e.,
\begin{equation}
IS_{i} =\max_{j} IS_{i, j}
\end{equation}
Then, the final interpretability score of a word embedding space, denoted by $IS$, can be defined as the average interpretability score across all dimensions, i.e.,
\begin{equation}\label{eq:interpretability}
 IS=\frac{1}{D}\sum_{i=1}^{D}IS_{i}
\end{equation}
%

\subsection{Intrinsic Evaluation}
 Word embedding algorithms such as word2vec and GloVe capture semantic and syntactic word relations to different extents. One approach to evaluating the performance of a word embedding method, is intrinsic evaluation which is typically performed on sets of word vectors in order to measure their overall success in maintaining the semantic and syntactic relation among the words, regardless of a specific downstream task. Intrinsic evaluation tasks are mostly faster and easier to carry out compared to extrinsic evaluation tasks which are discussed in the next section.

In order to measure word relations quantitatively, many benchmarks have been proposed for both medical and non-medical domains. These benchmarks consist of word pairs and their corresponding human-rated similarity and relatedness scores as references. While the scores are usually scaled to 10, there are a few exceptions to this convention. The word sets that are used for intrinsic evaluation in this paper are as follows:
\begin{itemize}
    \item SimLex-999 \cite{hill2015simlex}: This set  includes 999 word pairs comprising noun, verb, and adjective pairs scored by raters recruited from Amazon Mechanical Turk. SimLex-999 provides scores based on only similarity among the pairs and not relatedness or other associations.
    \item Bio-SimLex and Bio-SimVerb \cite{chiu2018bio}: These sets contain 988 noun pairs and 1000 verb pairs in medical domain, respectively. They are labeled by annotators with a biology background. Same as SimLex, the ratings are based on similarity alone.
    \item UMNSRS \cite{UMNSRS}: In contrast to the word sets mentioned above, UMNSRS  has two separate datasets for similarity and relatedness associations consisting of 566 and 588 word pairs in biomedical area, respectively. These words cover a diverse fields including drugs, medicines, disorders, etc.
\end{itemize}
To evaluate the performance of the word embeddings with respect to each reference set, we compute the cosine similarity between the vectors of each pair in the reference set. Once the similarity scores are obtained, their Spearman rank correlations with the scores given in the reference set are calculated. Higher values of correlation indicate better capturing of the word relations in the embedding.

\subsection{Extrinsic Evaluation}\label{Sec4.6}
Extrinsic evaluation is another approach to assessing the performance of
word embeddings. The idea is to evaluate the efficiency of the word vectors when used as input to a downstream (classification) task. The tasks considered in this paper for extrinsic evaluation are as follows:
\begin{itemize}
\item \emph{Polarity Classification (PC) and Factuality Classification (FC)}: We acquired the sentiment analysis dataset constructed in \cite{factuality} which is extracted from the patients' conversations on the medical forums of the MedHelp website\footnote{www.medhelp.org}. This dataset consists of a 3792 labeled sentences from patients suffering from food poisoning, crohn’s disease, and breast cancer. Each sentence is labeled in two different aspects. First, the factuality of the sentence that indicates what the comment is based on which is either,    ``Opinion", ``Fact", or ``Experience". The other label is based on the polarity of sentences, which is either ``Positive", ``Negative" or ``Neutral".

\item \emph{Question Classification (QC)}: This task and its accompanying dataset is inspired by the TREC dataset \cite{questionclassification} for question classification as a facilitating task for question answering. Specfically, we collected a set of questions asked by patients from the Medhelp medical forums. The dataset has roughly 2000 samples including 9 various conditions related to the digestive system such as Irritable bowel syndrome (IBS) and  Gastroesophageal Reflux Disease (GERD). Each class contains more than 200 questions and the task is to identify the corresponding condition from the question's text.
\end{itemize}

For both tasks, we use Support Vector Machine (SVM), Random Forest, Gradient Boosting, Passive Aggressive classifiers, Gaussian Naive Bayes, and the best performance in terms of \emph{accuracy} is reported. For further validation of the results, we apply ten-fold cross-validation and report the average accuracy for each task.

\section{Results and Discussion}\label{Sec5}
In this section, we provide the results for the experiments explained in the previous section. We begin with the results for the downstream tasks related to the intrinsic and extrinsic evaluation of the word embeddings.
\subsection{Downstream Tasks}
Table \ref{tab:intrinsic_res} summarizes the intrinsic evaluation of
the original embeddings (GloVe and Skip-gram) and their relevant sparse versions
through SPINE and SPOWV.
One important reason for looking into such comparison is to make sure that the expected interpretability of the sparse methods is achieved at either no or tolerable loss in preserving the similarity and relatedness captured by their original versions.
As it can be seen, sparse word vectors obtained by SPOWV have even outperformed the original vectors on several occasions. The sparse vectors obtained by SPINE however, show a slight loss of performance which seems nonetheless tolerable considering the high interpretability it provides, as shown in the next subsection.


As for the extrinsic evaluation, Table \ref{tab:extrinsic_res} summarizes the performance of the six embeddings under study, in terms of accuracy, for the aforementioned classification tasks, explained in Section~\ref{Sec4.6}.
As it can be observed, the classifiers benefit slightly from the sparsity of the vectors.

\begin{table*}[!t]
\caption{Word similarity and relatedness results}
\renewcommand{\arraystretch}{1.3}
\label{tab:intrinsic_res}
\begin{center}
{\small
\begin{tabular}{|c|c|c|c|c|c|}
\hline
\textbf{Vectors} & \textbf{SimLex} & \textbf{Bio-SimLex} & \textbf{Bio-SimVerb} & \textbf{UMNSRS-Sim} & \textbf{UMNSRS-Rel} \\ \hline\hline
\textbf{GloVe} & 0.29 & 0.66 & 0.53 & 0.56 & \textbf{0.58} \\
\textbf{SPINE GloVe} & 0.28 & 0.66 & 0.47 & 0.53 & 0.52 \\
\textbf{SPOWV GloVe} & 0.30 & 0.66 & \textbf{0.54} & 0.56 & 0.57 \\ \hline\hline
\textbf{Skip-gram} & 0.30 & \textbf{0.67} & 0.50 & 0.56 & 0.55 \\
\textbf{SPINE SG} & 0.25 & 0.60 & 0.45 & 0.44 & 0.43 \\
\textbf{SPOWV SG} & \textbf{0.31} & 0.65 & 0.48 & \textbf{0.60} & \textbf{0.58} \\ \hline
\end{tabular}}
\end{center}
\end{table*}
\begin{table}[!t]
\centering
\caption{Extrinsic evaluation results}
\renewcommand{\arraystretch}{1.2}
\label{tab:extrinsic_res}
{\small
\begin{tabular}{|c|ccc|c|}
\hline
\textbf{Vectors} & \multicolumn{1}{c|}{\textbf{\begin{tabular}[c]{@{}c@{}}PC\end{tabular}}} & \multicolumn{1}{c|}{\textbf{\begin{tabular}[c]{@{}c@{}}FC\end{tabular}}} & \textbf{\begin{tabular}[c]{@{}c@{}}QC\end{tabular}} & \textbf{Average} \\ \hline \hline
\textbf{GloVe} & 60.9 & 68.3 & 82.5 & 70.5 \\
\textbf{SPINE GloVe} & 61.3 & 68.0 & 82.8 & 70.7 \\
\textbf{SPOWV GloVe} & \textbf{62.2} & \textbf{69.2} & \textbf{82.9} & \textbf{71.4} \\ \hline\hline
\textbf{Skip-gram} & 61.2 & 68.4 & 82.9 & 70.8 \\
\textbf{SPINE SG} & 60.4 & 67.8 & 83.1 & 70.4 \\
\textbf{SPOWV SG} & \textbf{62.2} & \textbf{69.5} & \textbf{83.6} & \textbf{71.7} \\ \hline
\end{tabular}}
\end{table}

\subsection{Evaluation of Interpretability}
In this section, we provide the results  on evaluating the interpretability of the considered embeddings.  We begin with qualitative results.
\subsubsection{Qualitative Evaluation}


In terms of interpretability, we begin with an example which visualizes the obtained sparse vectors of words to provide an intuition of interpretability and sparsity from the distribution of values upon dimensions for each word vector.

Consider the following six words:
1) Melanoma, 2) Colorectal cancer,  3) Ewing's sarcoma\footnote{A type of cancer developed in bone or soft tissue.}, 4) Acetaminophen, 5) Aspirin, 6) Clopidogrel. The first three words belong to the class of cancer types, and the next three words are pharmaceutical drugs.

Figure \ref{fig-} shows the dimensions of the vector representations of these six words sorted by the average value across the three cancer type vectors.
As it can be seen, for both SPINE and SPOWV, the distribution of values over dimensions have the same pattern in the three types of cancer compared to the words that are medicines.\footnote{Note that SPOWV vectors contain negative values, while SPINE vectors are non-negative.}
A similar result is observed if the vectors are sorted by the average of the drug type vectors.

As we discussed earlier, for an interpretable embedding, we expected dimensions to represent specific semantic groups in the sense that the words active in an interpretable dimension form a semantically coherent group. To examine this, we ran the experiment explained in Section~\ref{inter_qual} , the results of which are shown in Table \ref{tab:qual_inter_res}. The medical terms used to carry out the experiment are
Asthma, Alzheimer, Insulin and NRAS\footnote{A class of genes that can become cancerous if mutated.}.

The results summarized in Table \ref{tab:qual_inter_res} suggest that the top words in the dominating dimension of each selected word are related to each other and the selected word, in the sparse representations, especially in SPINE vectors. As it can be seen, both SPOWV and SPINE vectors are shown to be much more interpretable than the original ones. For example, the top words in the dominating dimension of the vector representing \emph{Insulin} in the SPINE version of GloVe, are rosiglitazone, hyperinslinemia, IGF, hyperglycemia, and pioglitazone, which are all closely related to insulin and diabetes. This is while the original vectors fail to display such signs of interpretability. The same is true for the other words in this experiment.

\begin{figure*}[!t]\label{fig-}
\begin{minipage}{0.5\linewidth}
  \centering
  \centerline{\includegraphics[width=.97\linewidth]{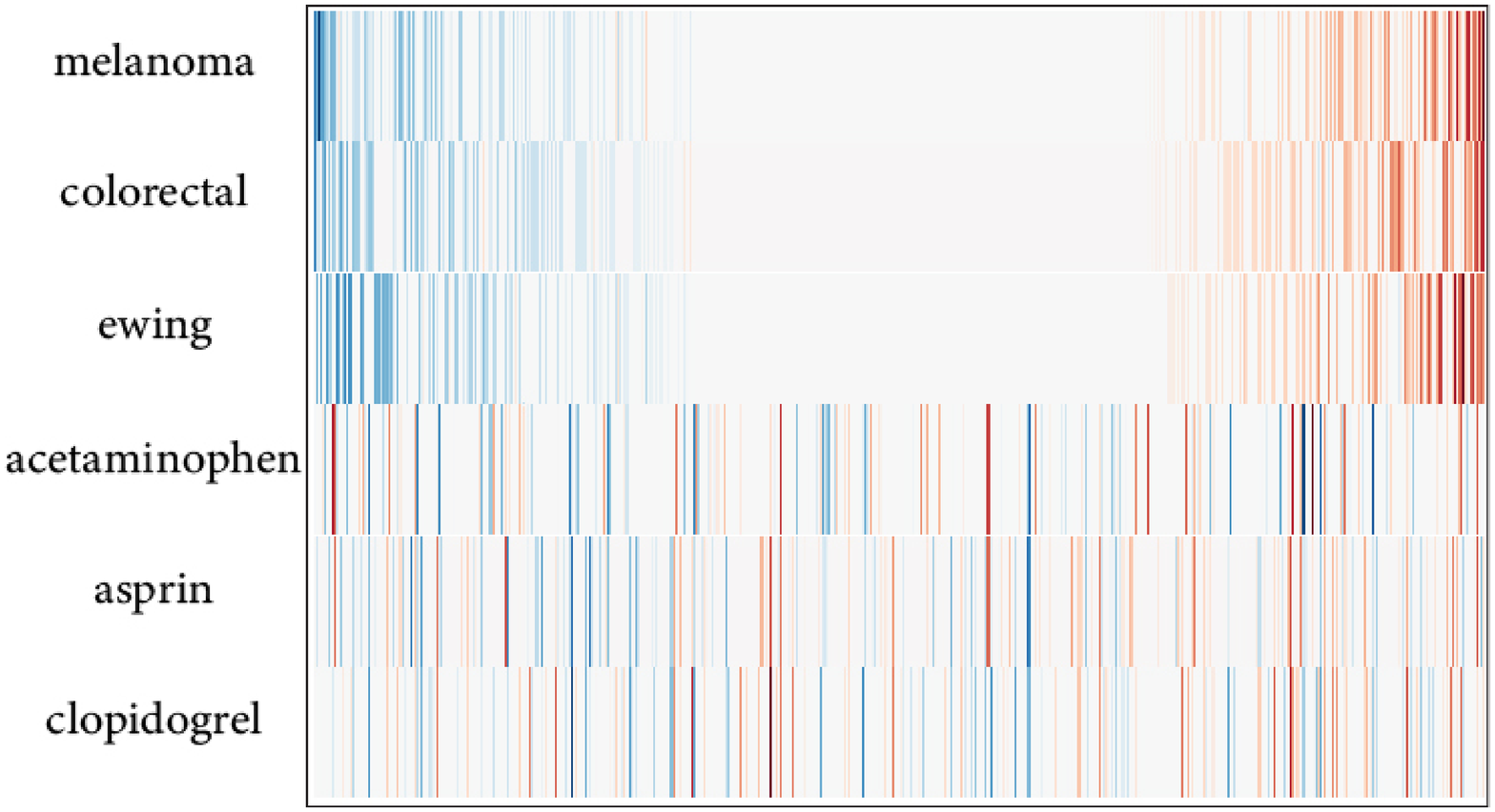}}
  \vspace{0.01cm}
\centerline{(a)}\medskip
\end{minipage}
\begin{minipage}{0.5\linewidth}
  \centering
  \centerline{\includegraphics[width=.97\linewidth]{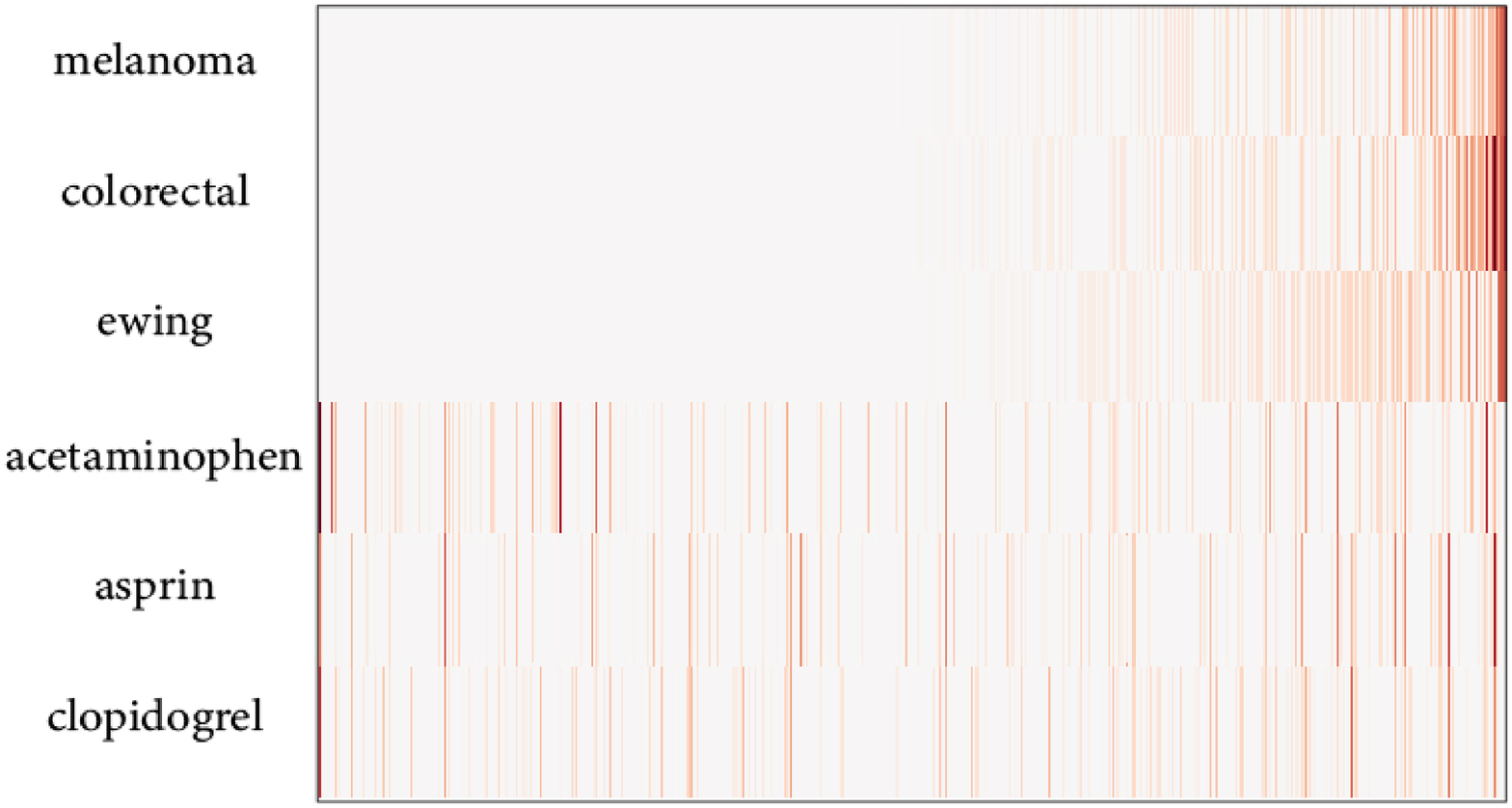}}
  \vspace{0.01cm}
\centerline{(b)}\medskip
\end{minipage}
\caption{Visualization of the dimensions sorted by the average value of the first 3 words (the cancer vectors), where the positive, negative and zero values are shown by the red, blue, and white lines, respectively. (a) SPOWV GloVe; (b) SPINE GloVe}\label{fig-}
\end{figure*}
\begin{table}[!t]
\caption{Qualitative evaluation of interpretability}
\renewcommand{\arraystretch}{1.3}
\label{tab:qual_inter_res}
\begin{center}
{\small
\begin{tabular}{cccc}
\hline
\multicolumn{1}{|c|}{\textbf{Concept}} & \multicolumn{1}{c|}{\textbf{GloVe}} & \multicolumn{1}{c|}{\textbf{SPINE GloVe}} & \multicolumn{1}{c|}{\textbf{SPOWV GloVe}} \\ \hline\hline
\multicolumn{1}{|c|}{\textbf{Asthma}} & \multicolumn{1}{c|}{\begin{tabular}[c]{@{}c@{}}polycystic, immunotherapy\\ mansoni, schistosome\\ chemoresistance\end{tabular}} & \multicolumn{1}{c|}{\begin{tabular}[c]{@{}c@{}}copd, comorbid, fibromyalgia, \\ rhinitis, debilitating\end{tabular}} & \multicolumn{1}{c|}{\begin{tabular}[c]{@{}c@{}}sputum, perennial, herb, \\ exocrine, peach\end{tabular}} \\ \hline
\multicolumn{1}{|c|}{\textbf{Alzheimer}} & \multicolumn{1}{c|}{\begin{tabular}[c]{@{}c@{}}untrained, mental, \\ illnesses, srh, youths\end{tabular}} & \multicolumn{1}{c|}{\begin{tabular}[c]{@{}c@{}}europathology,  neuroinflammation,\\  prion, amyloid, syn\end{tabular}} & \multicolumn{1}{c|}{\begin{tabular}[c]{@{}c@{}}parkinson, dementia, \\ falls, slips, levodopa\end{tabular}} \\ \hline
\multicolumn{1}{|c|}{\textbf{Insulin}} & \multicolumn{1}{c|}{\begin{tabular}[c]{@{}c@{}}crb, lithium, amiodarone,\\  gnrh, disagreement\end{tabular}} & \multicolumn{1}{c|}{\begin{tabular}[c]{@{}c@{}}rosiglitazone, hyperinsulinemia, \\ igf, hyperglycemia, pioglitazone\end{tabular}} & \multicolumn{1}{c|}{\begin{tabular}[c]{@{}c@{}}influencing, lobes. \\ macrovascular. abiotic. \\ hydatid\end{tabular}} \\ \hline
\multicolumn{1}{|c|}{\textbf{NRAS}} & \multicolumn{1}{c|}{\begin{tabular}[c]{@{}c@{}}oncogenes, upa cdx, \\ ceacam, trophoblast\end{tabular}} & \multicolumn{1}{c|}{\begin{tabular}[c]{@{}c@{}}tumours, kras, hnscc, \\ cancers, mutated\end{tabular}} & \multicolumn{1}{c|}{\begin{tabular}[c]{@{}c@{}}nras, pod, daf,\\  truncation, ontogenetic\end{tabular}} \\ \hline
 &  &  &  \\ \hline
\multicolumn{1}{|c|}{\textbf{Concept}} & \multicolumn{1}{c|}{\textbf{Skip-gram}} & \multicolumn{1}{c|}{\textbf{SPINE Skip-gram}} & \multicolumn{1}{c|}{\textbf{SPOWV Skip-gram}} \\ \hline\hline
\multicolumn{1}{|c|}{\textbf{Asthma}} & \multicolumn{1}{c|}{\begin{tabular}[c]{@{}c@{}}vhl, cdh, jak, \\ chromosomal. nonsense\end{tabular}} & \multicolumn{1}{c|}{\begin{tabular}[c]{@{}c@{}}bal, cf, airways, \\ balf, inhalation\end{tabular}} & \multicolumn{1}{c|}{\begin{tabular}[c]{@{}c@{}}interferon, virological, \\ isg, antiviral, virologic\end{tabular}} \\ \hline
\multicolumn{1}{|c|}{\textbf{Alzheimer}} & \multicolumn{1}{c|}{\begin{tabular}[c]{@{}c@{}}deacetylase, bioconductor, \\ microbe, html, idd\end{tabular}} & \multicolumn{1}{c|}{\begin{tabular}[c]{@{}c@{}}hippocampalad, brains\\ neuropathology,  neurodegeneration\\hippocampal\end{tabular}} & \multicolumn{1}{c|}{\begin{tabular}[c]{@{}c@{}}montreal, james, hads\\ edinburgh, gba\end{tabular}} \\ \hline
\multicolumn{1}{|c|}{\textbf{Insulin}} & \multicolumn{1}{c|}{\begin{tabular}[c]{@{}c@{}}dog, alarm, fox, \\ chimera, xenopus\end{tabular}} & \multicolumn{1}{c|}{\begin{tabular}[c]{@{}c@{}}insulin, glp, ins, \\ somatostatin, islets\end{tabular}} & \multicolumn{1}{c|}{\begin{tabular}[c]{@{}c@{}}glu, glycated, inh,\\  ins, freeze\end{tabular}} \\ \hline
\multicolumn{1}{|c|}{\textbf{NRAS}} & \multicolumn{1}{c|}{\begin{tabular}[c]{@{}c@{}}vhl, cdh, jak, \\ chromosomal, nonsense\end{tabular}} & \multicolumn{1}{c|}{\begin{tabular}[c]{@{}c@{}}pms, mutations, lynch, \\ pdgfra, germline\end{tabular}} & \multicolumn{1}{c|}{\begin{tabular}[c]{@{}c@{}}braf, thrombolysis, \\ codons, pik, idh\end{tabular}} \\ \hline
\end{tabular}}
\end{center}
\end{table}

\subsubsection{Quantitative Evaluation}

We finally measure the interpretability of each method by applying the proposed quantitative scoring approach explained in Section \ref{inter_quant}. The results are provided in Table \ref{tab:quant_inter_res} which shows the average interpretability score across all dimensions for each method. The results agree with those obtained by qualitative evaluation.
As it can be seen, although SPOWV vectors were seen to perform better in downstream tasks, vectors trained by SPINE have better interpretability in medical domain.

\begin{table}[!t]
\caption{Quantitative evaluation of interpretability}
\renewcommand{\arraystretch}{1.3}
\label{tab:quant_inter_res}
\begin{center}
{\small
\begin{tabular}{cccc}
\hline
\multicolumn{1}{|c|}{\textbf{GloVe}} & \multicolumn{1}{c|}{\textbf{Original}} & \multicolumn{1}{c|}{\textbf{SPINE}} & \multicolumn{1}{c|}{\textbf{SPOWV}} \\ \hline
\multicolumn{1}{|c|}{\textbf{Interpretability Score}} & \multicolumn{1}{c|}{8.6} & \multicolumn{1}{c|}{\textbf{18.1}} & \multicolumn{1}{c|}{9.9} \\ \hline
 &  &  &  \\ \hline
\multicolumn{1}{|c|}{\textbf{Skip-gram}} & \multicolumn{1}{c|}{\textbf{Original}} & \multicolumn{1}{c|}{\textbf{SPINE}} & \multicolumn{1}{c|}{\textbf{SPOWV}} \\ \hline
\multicolumn{1}{|c|}{\textbf{Interpretability Score}} & \multicolumn{1}{c|}{9.6} & \multicolumn{1}{c|}{\textbf{16.7}} & \multicolumn{1}{c|}{12.3} \\ \hline
\end{tabular}}
\end{center}
\end{table}

\section{Conclusion and Future Work}\label{Sec6}

In the context of NLP for medical domain, we compared the interpretability of the state-of-the-art word embedding methods with the two most nominated sparse interpretable word vector embeddings.
We proposed a novel approach for quantifying the interpretability of the word vectors based on category theory. The approach was carried out by a dataset of semantic coherent groupings of medical terminologies that we proposed in this regard. The sparse word vectors trained showed much more interpretability, without downgrading the performance of their original vectors in downstream tasks.

For future work, several directions remain open. Since SPINE and SPOWV are sparse, they are computationally efficient. Therefore, areas such as energy and timing analysis can be further investigated.  Moreover, as the results suggest, although SPINE models improved the interpretability most desirably, they suffered from a marginal loss in preserving similarity and word associations, compared to SPOWV. Adding an extension to the cost function of the SPINE to further maintain the similarity can be another direction for future studies.

\bibliography{acl2020}
\bibliographystyle{acl_natbib}
\end{document}